\documentclass[letterpaper, 10 pt, conference]{ieeeconf}  

\IEEEoverridecommandlockouts                              

\usepackage{xcolor}
\usepackage{graphicx}
\usepackage{booktabs}
\usepackage{multirow}
\usepackage[ruled,vlined]{algorithm2e}
\usepackage{amsmath}

\usepackage{todonotes}

\title{\LARGE \bf
Follow the Signs: Using Textual Cues and LLMs to \\ Guide Efficient Robot Navigation

}

\author{Jing Cao, Nishanth Kumar*, and Aidan Curtis* 
\\
MIT CSAIL
\thanks{$^{*}$Equal advising. Primary email contact: \tt\small jingcao@mit.edu.}}

\begin{document}

\maketitle
\thispagestyle{empty}
\pagestyle{empty}

\begin{abstract}
    Autonomous navigation in unfamiliar environments often relies on geometric mapping and planning strategies that overlook rich semantic cues such as signs, room numbers, and textual labels. We propose a novel semantic navigation framework that leverages large language models (LLMs) to infer patterns from partial observations and predict regions where the goal is most likely located. Our method combines local perceptual inputs with frontier-based exploration and periodic LLM queries, which extract symbolic patterns (e.g., room numbering schemes and building layout structures) and update a confidence grid used to guide exploration. This enables robots to move efficiently toward goal locations labeled with textual identifiers (e.g., “room 8”) even before direct observation. We demonstrate that this approach enables more efficient navigation in sparse, partially observable grid environments by exploiting symbolic patterns. Experiments across environments modeled after real floor plans show that our approach consistently achieves near-optimal paths and outperforms baselines by over 25\% in Success weighted by Path Length.
\end{abstract}

\label{sec:introduction}
\section{Introduction}

A key challenge for autonomous robots, especially in applications such as search-and-rescue or mobile manipulation, is to navigate efficiently within unfamiliar indoor environments.
Traditional approaches rely on a two-step pipeline: they first explore the environment to create a geometric map \cite{ebadi2020lamplargescaleautonomousmapping, blöchliger2018topomaptopologicalmappingnavigation}, and then plan in this map to reach specific targets \cite{ctx23201582960006761}.
While this strategy is effective for generating collision-free trajectories, it can be extremely inefficient due to the time taken for initial exploration, especially in large buildings.

Humans intuitively navigate unfamiliar environments effectively by leveraging semantic cues, such as textual labels, signs, and room numbers.
These cues enable them to make informed decisions about their paths even before they directly observe their destinations. 
For example, encountering a sign that reads ``Rooms 10-20 $\rightarrow$" is often enough to determine a possible path to Room 12 without direct observation. 

In this work, we seek to develop a navigation approach that exploits rich semantic cues to enable efficient goal-reaching in partially-observable structured environments.
Recent work has demonstrated that large language models (LLMs) are able to interpret semantic information to improve navigation in unstructured household environments \cite{shah2023lfg, shah2025foresightnavlearningsceneimagination, anderson2018visionandlanguagenavigationinterpretingvisuallygrounded}. 
These methods leverage semantic regularities grounded in object categories (e.g., fridges in kitchens, beds in bedrooms), enabling large language models (LLMs) or vision–language models (VLMs) to associate scene context with likely goal locations.

However, such object-centric reasoning does not transfer well to large structured environments like offices or hotels.
In homes, semantic regularities are grounded in object categories (e.g., fridges in kitchens, beds in bedrooms), which LLMs and VLMs can exploit to infer likely goal locations.
By contrast, structured environments are organized around textual cues such as room numbers, directional signage, and numbering schemes, which require reasoning over abstract spatial patterns rather than object co-occurrence.
Existing approaches that treat LLMs as end-to-end planners struggle in this setting because LLMs are not inherently spatially grounded, they often hallucinate implausible navigation strategies and exhibit brittle behavior, particularly under partial observability or perception noise.

We choose to tackle these challenges by combining LLMs with classical navigation and exploration algorithms. 
The key idea is to use LLMs to \textit{hypothesize} a goal region from semantic cues in the current observation and from a partial map constructed with accumulated observations during exploration.
Intuitively, our approach combines the complementary strengths of both paradigms: LLMs provide semantic reasoning while classical exploration enables full coverage and reliability in large structured spaces. 

In experiments, we find that our approach consistently outperforms methods that rely soley on classical navigation or LLM-based planning. 
Evaluated across seven environments of varying complexity, our method achieves near-optimal paths and the highest Sucess weighted by Path Length (SPL) anong all baselines. Overall, our method attains an SPL of $0.745$, compared to $0.596$ for Frontier Exploration, $0.236$ for NavGPT, and $0.160$ for LLM-Only,  while producing paths more than 40\% shorter than alternatives.

\section{Related Works}
\label{sec:related}

\begin{figure*}[t]
    \includegraphics[width=\linewidth]{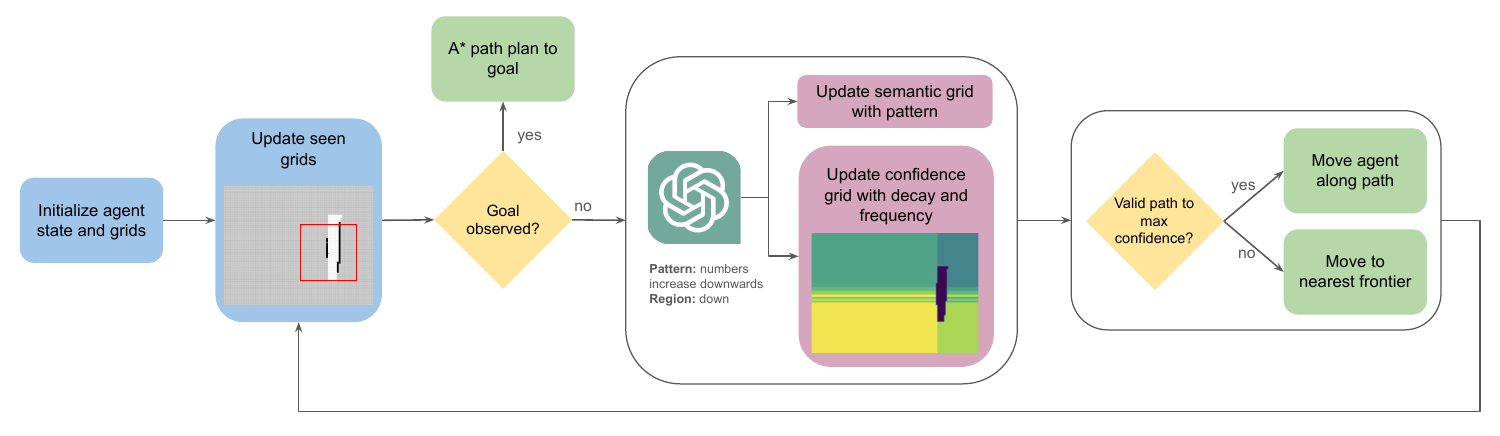}
    \centering
    \caption{Overview of navigation pipeline. 
    At each timestep, the agent updates its seen grids from local observations. 
    If the goal is directly visible, it plans a shortest path to it using A*. 
    Otherwise, the LLM predicts a likely goal region, the confidence grid is updated, 
    and the agent plans a path either toward the highest-confidence region or, if the confidence grid is uniform, toward the nearest unexplored frontier. 
    This loop repeats until the goal is found or the environment is fully explored.
    }
    \label{fig:method-overview}
\end{figure*}

\textbf{Scene graphs:} Scene graphs provide a structured semantic representation of an environment, often used to support spatial reasoning. Systems like Hydra~\cite{hughes2022hydrarealtimespatialperception} construct 3D scene graphs in real-time for localization and object tracking. 
Recent works have grounded LLMs in scene graphs to scale up task planning \cite{rana2023sayplangroundinglargelanguage} or to enable dynamic goal-directed planning in unseen environments \cite{rajvanshi2024saynav}. 
However, scene graphs typically require a constructed map ahead of time and do not leverage the semantics of the environment, such as room numbering patterns or textual annotations that can guide exploration. By contrast, our method infers these semantic regularities directly from partial observations, enabling flexible navigation in sparse and unstructured settings where no prior map is available.

\textbf{Planning with LLMs:} LLMs have been successfully applied to robot planning, where they translate natural language commands into symbolic plans, generate motion plans that align with environmental constraints, and decompose abstract tasks into executable action sequences without additional training \cite{xie2023translatingnaturallanguageplanning, Lin_2023, huang2023groundeddecodingguidingtext, curtis2024trustproc3ssolvinglonghorizon, codeaspolicies2022, pmlr-v205-ichter23a}. These methods demonstrate that pre-trained language models have high-level world knowledge that can be leveraged for task execution. While strong in semantic understanding and symbolic task decomposition, these methods mainly focus on manipulation tasks under fully observable conditions. 
Our work is focused on large-scale navigation under partial observability. LLMs have also demonstrated limitations in spatial reasoning, making their standalone use in navigation tasks insufficient \cite{bubeck2023sparksartificialgeneralintelligence, yamada2024evaluating}.

\textbf{Vision and language navigation:} Vision and language navigation systems have typically relied on predefined mappings between semantic cues and agent actions. Some approaches have used symbolic reasoning frameworks that depend on manually annotated signs and rules to guide exploration \cite{7139313}, while others have applied visual place recognition techniques that detect and interpret scene text for localization \cite{10.3389/frobt.2024.1424883}. Natural language interfaces have also been developed to parse user commands into structured action sequences or goals \cite{Tellex_Kollar_Dickerson_Walter_Banerjee_Teller_Roy_2011}, and systems have been designed to follow visually-grounded instructions based on image-text alignments \cite{anderson2018visionandlanguagenavigationinterpretingvisuallygrounded}. More sophisticated models have fused visual inputs with textual annotations into joint embeddings or topological maps to support navigation \cite{huang2023visuallanguagemapsrobot, majumdar2020improvingvisionandlanguagenavigationimagetext}. However, these models often fail to capture higher-level structural patterns in environments -- such as the spatial ordering of room numbers or the hierarchical layout of buildings -- which are essential for efficient semantic reasoning of a new environment.

\textbf{Navigation with LLMs:} Recent work has sought to use LLMs in navigation tasks by either constructing foundation models designed for navigation \cite{shah2023vintfoundationmodelvisual, hirose2025learning} or by leveraging LLMs as tools for semantic reasoning and planning \cite{zhu2025strivestructuredrepresentationintegrating, barkley2025semanticintelligenceintegratinggpt4, huang2023groundeddecodingguidingtext}. These approaches typically involve training or prompting LLMs to predict goal locations, semantic layouts, or action sequences based on partial visual input or textual priors. Several works have introduced structured representations or reannotation techniques to enhance spatial generalization and exploration efficiency. However, several of these systems rely exclusively on the LLM's output to drive navigation \cite{10.1609/aaai.v38i7.28597, shah2022lmnavroboticnavigationlarge}, which can lead to inefficiencies or unreliable exploration in large, structured environments. In contrast, our approach builds spatial structure by augmenting LLM inference with classical navigation algorithms like frontier exploration and A* path planning to ensure full coverage of the environment. Furthermore, most existing works focus on household environments while we target semantically structured environments like office spaces, where spatial regularities (e.g., room numbering schemes) can be leveraged by the LLM to guide planning more effectively.

\section{Problem Formulation}
\label{sec:problem_formulation}

\subsection{Environment and observation model}
\begin{figure*}[!h]
    \includegraphics[width=\linewidth]{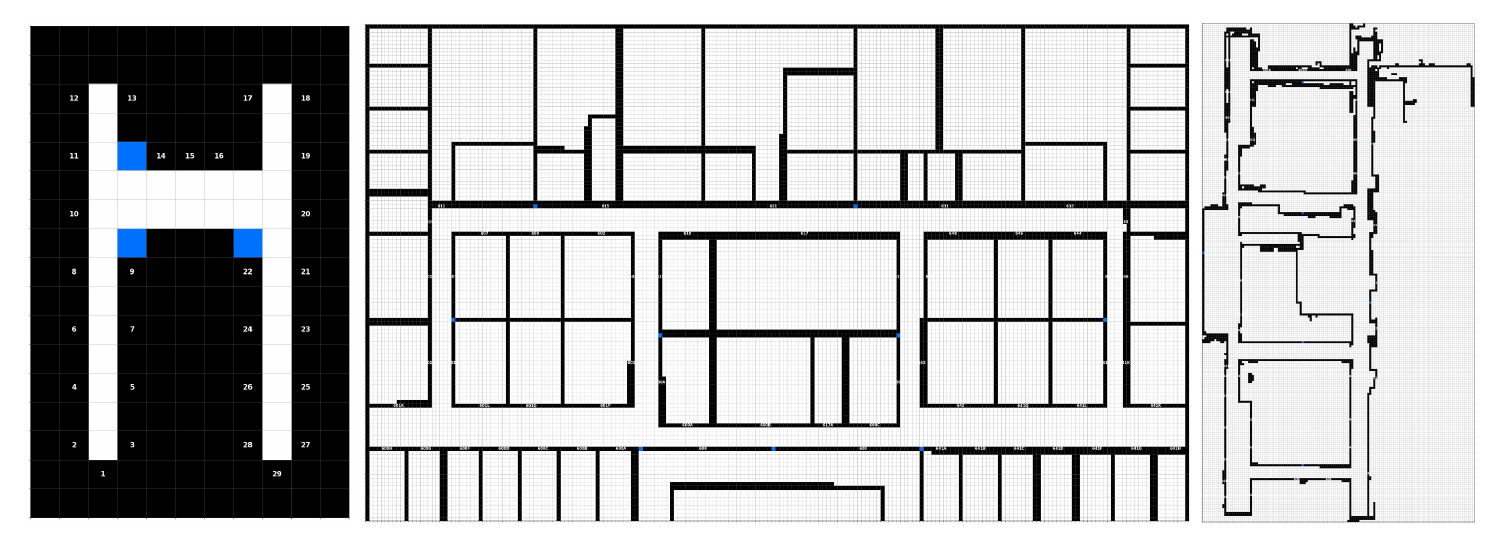}
    \centering
    \caption{Examples of grid environments. Black and white cells denote occupancy grid values, with black indicating obstacles and white indicating free space. Blue cells represent the locations of signs. Left: Small H-Shape environment; Middle: Large Offices environment; Right: Noisy Polycam environment.}
    \label{fig:env-examples}
\end{figure*}

We elect to model navigable environments as 2D grid worlds consisting of two primary components: an occupancy grid and a semantic grid. 
The binary occupancy grid defines structure and traversability. The corresponding semantic grid contains relevant features for describing the objects and their properties at that cell. 
This may contain discrete labels from a predefined set such as free space, walls, or doors as well as optional free-form attributes such as room number or sign text. 
The occupancy grid and semantic grid are correlated such that free spaces in the occupancy grid are labeled as free space in the semantic grid, and occupied spaces in the occupancy grid are labeled as either walls or doors in the semantic grid. 
Figure~\ref{fig:env-examples} illustrates examples of small, large, and noisy environments represented in this format. 

Additionally, at each timestep, the agent perceives a local $k \times k$ window centered at its current position, corresponding to its field of view, that reveals cells in the occupancy grid and semantic grid.

\subsection{POMDP formulation}
Our objective is to design an approach that takes as input a natural language goal command, leverages environmental cues to explore toward that goal, and outputs low-level commands to control the agent. Towards this, we formulate the underlying problem as a partially-observable Markov decision process (POMDP) defined on a fixed ground-truth map. A POMDP is defined by the tuple $\langle{\mathcal{S}, \mathcal{O}, \mathcal{A}, \mathcal{T}, \mathcal{R}, \mathcal{Z}, \gamma}\rangle$. 

\paragraph{State space $\mathcal{S}$}
We consider the state space to be the space of possible occupancy grids corresponding to indoor 2D floorplans along with a dictionary that maps grid cells to textual or semantic information located within that grid cell such as signs, room numbers, and object or landmark descriptions. 
Note that the size of the state space is unbounded because each cell can contain any arbitrary strings.

\paragraph{Observation space $\mathcal{O}$}
Observations consist of local $k \times k$ windows around the agent. Each observation includes both structural information (free space, walls, doors) and any textual annotations (room labels, sign text) contained in those cells.

\paragraph{Action space $\mathcal{A}$}
The action space consists of four primitive navigation moves: up, down, right, left.

\paragraph{Transition model $\mathcal{T}$}
The transition model is a deterministic single-cell movement of the agent in the direction of the action unless that movement is blocked by a solid object (e.g. a wall). 

\paragraph{Reward function $\mathcal{R}$}
The reward function assigns a penalty of $-1$ at every cell that's not the goal and $0$ at the goal cell containing the target room number. 
Thus, the agent obtains the highest possible reward from reaching the goal as quickly as possible.

\paragraph{Observation model $\mathcal{Z}$}
Lastly, the observation model maps the full state to the agent’s local field of view, resolving text and semantic labels associated with signs and objects into string-based representations corresponding to their grid cell location.

Classical POMDP solvers assume explicit, queryable representations of the state, observation, transition, and reward components and often plan directly in belief space, that is, distributions over states. In our setting this assumption breaks down. The state space is open-ended, and observations carry rich semantics. Directional signage, numbering schemes, and landmarks impose long-range, nonlocal constraints that are hard to encode, so the resulting belief updates are not well captured by a fixed local generative model.

Rather than constructing a full, explicit belief and observation model, we adopt the organizing principles of belief-space planning while keeping the heavy machinery implicit. We build a compact, belief-like representation that a language model can reason over, allowing semantic cues to guide planning without an explicit observation model, a hand-specified state distribution, or an enumerated belief.

\section{Method}
\label{sec:method}

We propose an LLM-guided belief update mechanism. 
At each timestep, the agent fuses local occupancy and semantic observations into partial visibility grids, uses an LLM to infer higher-level patterns, and then uses these inferences to update a confidence grid representing its belief over potential goal regions. 
This confidence grid, combined with frontier exploration, guides the agent’s control policy, enabling it to balance semantic reasoning with geometric exploration. 
Importantly, since the policy defaults to frontier exploration when semantic guidance is uncertain, and frontier exploration is known to be probabilistically complete under standard assumptions~\cite{1570713}, our overall algorithm inherits this property. 

This section describes the components of our approach: the perceptual updating, confidence grid construction, LLM-based semantic inference, and the resulting planning strategy.

\subsection{Goal extraction}
We begin by parsing the natural language command into a structured target. Given an instruction such as “Go to Room 621”, an LLM is prompted to extract the specific goal identifier, in this case the string “621”.

\subsection{Agent perception}
The agent maintains an internal representation of its environment using two visibility grids: one for occupancy (\texttt{SeenOccupancyGrid}) and one for semantics (\texttt{SeenSemanticGrid}). The \texttt{SeenOccupancyGrid} is initialized with $-1$ values to denote unobserved cells, while the \texttt{SeenSemanticGrid} is initialized with empty dictionaries in all cells. These visibility grids aid in updating agent’s belief state, recording what parts of the environment have been explored and what semantic information has been observed so far.  

\begin{figure*}[!h]
    \includegraphics[width=\linewidth]{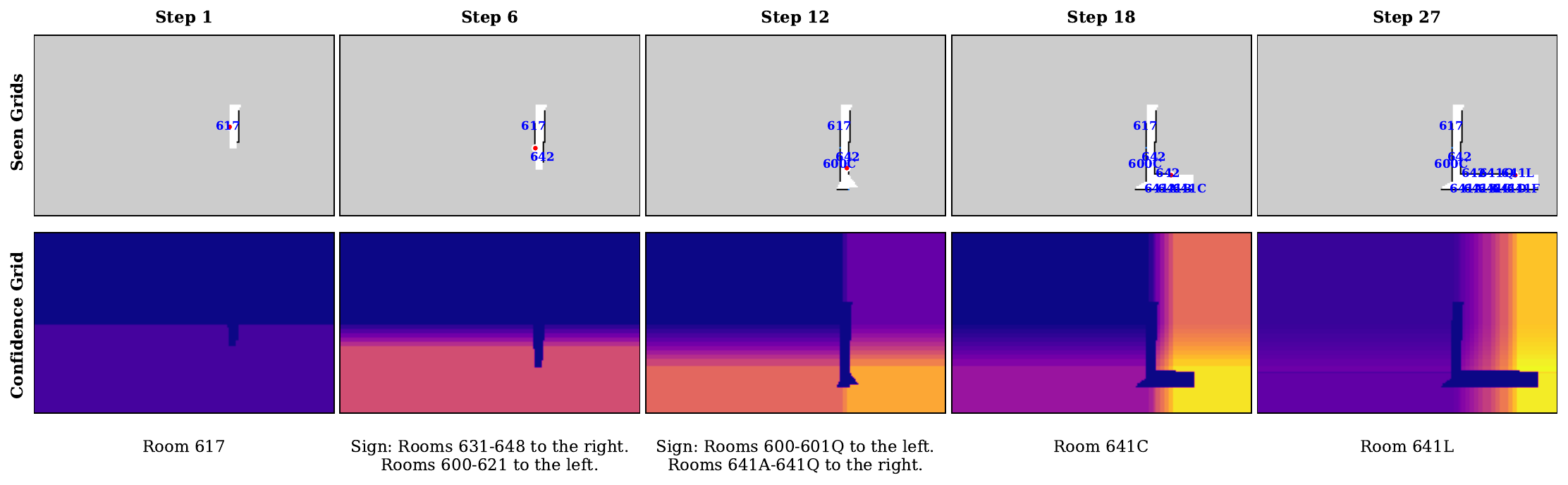}
    \centering
    \caption{Evolution of the agent’s grids over time for the goal of ``find room 641L''. The top row shows a fusion of the seen semantic and seen occupancy grids with blue dots representing signs and blue text representing room numbers. The bottom row shows the confidence grid with higher confidence indicating likely target locations.
    The grid is initialized with all zeros. As the LLM predicts likely goal directions, confidence values increase in corresponding regions, while explored cells inconsistent with the goal are reset to zero. Over time, repeated predictions with exponential decay lead to smooth gradients of confidence that guide exploration toward the most likely goal region (Step 27).}
    \label{fig:conf-grid}
\end{figure*}

At each timestep, the agent perceives a local $k \times k$ window centered at its current position, corresponding to its field of view. The contents of this window are used to update the visibility grids, incrementally expanding the agent’s knowledge of the structure and semantics of the environment. After each update, the LLM infers higher-level patterns from the observed semantics (e.g., numbering schemes, directional cues) and annotates the \texttt{SeenSemanticGrid} accordingly. Figure~\ref{fig:conf-grid} illustrates how these partial observations accumulate to form a progressively richer internal world model.

\subsection{Belief state modeling with confidence grid}
To maintain a probabilistic belief over the entire environment, we introduce a confidence grid. The confidence grid acts as a belief heatmap over where the goal likely is. 

The grid is initialized with zeros and updated after every LLM prediction (right, left, up, down). Based on the LLM's region prediction, all cells to the right, left, upwards, or downwards of the agent have their values increased by 1, denoting that the goal is more likely to be in that direction. To prevent older predictions from dominating, we apply an exponential decay factor $\alpha \in (0, 1)$ at each timestep so that recent predictions carry higher weight.

As the agent explores, the confidence grid updates to reflect the true state of the environment. For example, if the confidence grid shows that a cell is likely to be the goal but the agent has already explored the cell and knows that it's not the goal, that cell in the confidence grid will be set back to 0.

\subsection{LLM prediction}
At each timestep, the LLM is provided with the current \texttt{SeenOccupancyGrid}, \texttt{SeenSemanticGrid}, and the agent’s position. It outputs (i) a description of its reasoning process, (ii) semantic patterns inferred from the environment (e.g., room numbers increasing to the right, odd/even separation across hallways), and (iii) a predicted region $r$ relative to the agent where the goal is likely located. These predictions provide semantic guidance that can bias exploration toward likely goal regions.  

\subsection{Control loop}
Algorithm~\ref{alg:nav} summarizes the navigation procedure.  At each timestep, the agent must decide where to move next based on its partially observed belief of the environment and the evolving confidence grid. The algorithm maintains three internal representations: the \texttt{SeenOccGrid}, which stores partial observations of traversable space and obstacles, the \texttt{SeenSemGrid}, which stores semantic annotations such as room numbers or signs, and the confidence grid $C$, which tracks the belief distribution over where the goal is most likely to be located, with predictions decayed over time using a factor $\alpha$ to ensure older information is discounted.

The control policy proceeds as follows. First, the agent checks whether the goal has been directly observed in the \texttt{SeenSemGrid}. If so, the shortest collision-free path is computed with A* over the \texttt{SeenOccGrid} and executed immediately. If the goal is not visible, the agent queries the LLM with the current \texttt{SeenOccGrid}, \texttt{SeenSemGrid}, and position. The LLM predicts a relative region $r$ in which the goal is likely to be. The confidence grid $C$ is then updated by applying exponential decay $C \gets \alpha \cdot C$ and incrementing the cells corresponding to region $r$. Cells that have already been explored and ruled out as containing the goal are reset to zero, thereby preventing invalid hypotheses from persisting.

The centroid of the highest-confidence region is selected as the next subgoal, and A* is used to plan a path toward it. If the centroid coincides with the agent’s current position, the agent instead falls back to frontier exploration by selecting the nearest unexplored frontier (free cells adjacent to unknown cells) as the subgoal. In both cases, a collision-free path is generated with A*, and the agent executes the first step along this path. This process repeats until the goal is reached or the maximum horizon $T$ is exceeded.

    
        
        


\begin{algorithm}[t]
\caption{Our method}
\label{alg:nav}
\small
Initialize $SeenOccGrid$, $SeenSemGrid$, $C$\;
steps $\gets 0$\;

\While{steps $< T$}{
    $obs \gets$ local $k \times k$ window around agent\;
    Update $SeenOccGrid$, $SeenSemGrid$ with $obs$\;

    \If{goal $\in SeenSemGrid$}{
        $path \gets$ A*(agent, goal, $SeenOccGrid$)\;
        \Return $path$\;
    }

    $\text{pred} \gets$ LLM($SeenOccGrid$, $SeenSemGrid$, agent\_pos)\;
    $C \gets \alpha \cdot C$\;
    $C[r] \gets C[\text{pred}] + 1$\;

    \For{cell $c$}{
        \If{explored($c$) \textbf{and} $c \neq$ goal}{
            $C[c] \gets 0$\;
        }
        
    }

    $subgoal \gets$ centroid($\arg\max C$)\;

    \If{$subgoal = agent\_pos$}{
        $frontier \gets$ nearest\_frontier($SeenOccGrid$)\;
        $subgoal \gets frontier$\;
    }

    $path \gets$ A*(agent, subgoal, $SeenOccGrid$)\;
    Execute first action in $path$\;
}
\end{algorithm}

\begin{table*}[ht]
\centering
\caption{SPL and Success rate (mean $\pm$ std, 2 d.p.)}
\begin{tabular}{llcccccccc}
\toprule
Group & Env. & \multicolumn{2}{c}{NavGPT} & \multicolumn{2}{c}{LLM Only} & \multicolumn{2}{c}{Frontier Exploration} & \multicolumn{2}{c}{Ours} \\
\cmidrule(lr){3-4}\cmidrule(lr){5-6}\cmidrule(lr){7-8}\cmidrule(lr){9-10}
 &  & SPL & SR & SPL & SR & SPL & SR & SPL & SR \\
\midrule
\multirow{4}{*}{Small}
 & H-Shape      & 0.21 $\pm$ 0.30 & 0.60 $\pm$ 0.49 & 0.40 $\pm$ 0.35 & 0.70 $\pm$ 0.46 & 0.69 $\pm$ 0.33 & \textbf{1.00 $\pm$ 0.00} & \textbf{0.83 $\pm$ 0.29} & \textbf{1.00 $\pm$ 0.00} \\
 & Hallways  & 0.26 $\pm$ 0.29 & 0.90 $\pm$ 0.30 & 0.48 $\pm$ 0.34 & 0.70 $\pm$ 0.46 & 0.74 $\pm$ 0.32 & \textbf{1.00 $\pm$ 0.00} & \textbf{0.92 $\pm$ 0.19} & \textbf{1.00 $\pm$ 0.00} \\
 & Plaza        & 0.38 $\pm$ 0.30 & 0.90 $\pm$ 0.30 & 0.09 $\pm$ 0.17 & 0.40 $\pm$ 0.49 & 0.62 $\pm$ 0.31 & \textbf{1.00 $\pm$ 0.00} & \textbf{0.76 $\pm$ 0.32} & 0.90 $\pm$ 0.30 \\
 & \textbf{Overall} & 0.28 $\pm$ 0.29 & 0.80 $\pm$ 0.40 & 0.32 $\pm$ 0.29 & 0.60 $\pm$ 0.49 & 0.68 $\pm$ 0.32 & \textbf{1.00 $\pm$ 0.00} & \textbf{0.83 $\pm$ 0.27} & 0.97 $\pm$ 0.18 \\
\midrule
\multirow{4}{*}{Large}
 & H-Shape      & 0.10 $\pm$ 0.16 & 0.10 $\pm$ 0.30 & 0.05 $\pm$ 0.12 & 0.20 $\pm$ 0.40 & 0.74 $\pm$ 0.29 & \textbf{1.00 $\pm$ 0.00} & \textbf{0.82 $\pm$ 0.25} & \textbf{1.00 $\pm$ 0.00} \\
 & L-Shape       & 0.29 $\pm$ 0.30 & 0.50 $\pm$ 0.50 & 0.09 $\pm$ 0.18 & 0.30 $\pm$ 0.46 & 0.64 $\pm$ 0.32 & \textbf{1.00 $\pm$ 0.00} & \textbf{0.87 $\pm$ 0.22} & \textbf{1.00 $\pm$ 0.00} \\
 & Offices & 0.20 $\pm$ 0.27 & 0.40 $\pm$ 0.49 & 0.00 $\pm$ 0.00 & 0.00 $\pm$ 0.00 & \textbf{0.56 $\pm$ 0.34} & \textbf{1.00 $\pm$ 0.00} & 0.53 $\pm$ 0.31 & 0.90 $\pm$ 0.30 \\
 & \textbf{Overall} & 0.19 $\pm$ 0.24 & 0.33 $\pm$ 0.47 & 0.05 $\pm$ 0.10 & 0.17 $\pm$ 0.37 & 0.65 $\pm$ 0.32 & \textbf{1.00 $\pm$ 0.00} & \textbf{0.74 $\pm$ 0.26} & 0.97 $\pm$ 0.18 \\
\midrule
\multirow{2}{*}{Noisy}
 & Polycam       & 0.15 $\pm$ 0.27 & 0.20 $\pm$ 0.40 & 0.03 $\pm$ 0.10 & 0.20 $\pm$ 0.40 & 0.46 $\pm$ 0.33 & \textbf{1.00 $\pm$ 0.00} & \textbf{0.66 $\pm$ 0.31} & 0.80 $\pm$ 0.40 \\
 & \textbf{Overall} & 0.15 $\pm$ 0.27 & 0.20 $\pm$ 0.40 & 0.03 $\pm$ 0.10 & 0.20 $\pm$ 0.40 & 0.46 $\pm$ 0.33 & \textbf{1.00 $\pm$ 0.00} & \textbf{0.66 $\pm$ 0.31} & 0.80 $\pm$ 0.40 \\
\bottomrule
\end{tabular}
\label{tab:combined_spl_sr}
\end{table*}

\section{Experiments}
\label{sec:experiments}

\subsection{Experimental setup}

\textbf{Environments}
We evaluate our method across 7 different environments of varying complexity: 3 small synthetic environments, 3 large environments based on real floor plans, and 1 noisy environment constructed from a real-world Polycam scan of a building floor \cite{polycam}. 
Each environment is represented as a discrete grid through which the agent navigates.   

The grid sizes of each of the environments given by $(\text{rows}, \text{columns})$ is as follows:
\begin{enumerate}
    \item Small H-Shape: $(11, 7)$
    \item Small Hallways: $(7, 11)$
    \item Small Plaza: $(13, 7)$
    \item Large H-Shape: $(132, 122)$
    \item Large L-Shape: $(93, 244)$
    \item Large Offices: $(127, 211)$
    \item Noisy Polycam: $(251, 137)$
\end{enumerate}

Most environments contained purely numeric room identifiers, but the Large Offices and Noisy Polycam environments also contained alphanumeric identifiers (e.g., 641A).

\textbf{Approaches and Baselines}
We compare our method against the most related baseline from the literature, along with two ablations:
\begin{enumerate}
    \item NavGPT 
    \item LLM Only
    \item Frontier Exploration
\end{enumerate}

NavGPT is a vision-and-language navigation (VLN) agent that treats navigation as a text-based reasoning task powered by large language models. 
At each step, visual inputs are converted into natural language, and the LLM receives these scene descriptions together with the navigation instruction and a summarized history of past observations, reasoning, and actions. 
The model outputs both a reasoning trace and a next action, the action is executed in the environment, the history is updated, and the process continues~\cite{10.1609/aaai.v38i7.28597}. 
In our implementation, we directly provide the LLM with text descriptions of its observations.

The LLM Only baseline treats navigation as step-by-step action selection without global planning. 
At each timestep, the agent updates its partially observed occupancy and semantic grids. 
The LLM is provided the goal, current pose, and the seen occupancy and semantic grids. 
The LLM outputs the next action and the cycle repeats.

The Frontier Exploration baseline drives the agent towards the boundary between explored and unexplored space. 
At each step, the agent updates its seen occupancy and semantic grids. 
If the goal is visible in the semantic grid, the agent executes a final A* path to the goal. 
Otherwise, it samples one frontier and advances one step along an A* path through currently known free space. 

\textbf{Evaluation Criteria}
Performance is measured using two metrics. Success weighted by Path Length (SPL) quantifies path efficiency relative to the most optimal path and is defined as
    \begin{equation}
        \text{SPL} = \frac{1}{N} \sum_{i=1}^{N} S_i \cdot \frac{L_i}{\max(P_i, L_i)}
    \end{equation}
    
where $S_i$ is the success indicator (1 if the goal is reached, 0 otherwise), $L_i$ is the shortest path length, and $P_i$ is the actual path length taken by the agent in episode $i$. Success Rate (SR) measures the fraction of episodes in which the agent reaches the goal within the timeout.

\subsection{Results and analysis}
Table~\ref{tab:combined_spl_sr} displays our raw quantitative results. Bolded entries represent the best performing methods in each environment and group. The success rates less than 1 in our method are due to timeouts.

In our small environments (clean, synthetic layouts) our method achieves the best performance across all metrics. Frontier Exploration consistently succeeds but produces longer trajectories because it does not leverage semantic cues. LLM Only takes longer and fails frequently due to poor spatial reasoning and often proposes invalid actions (e.g., colliding with obstacles). NavGPT underperforms because its history buffer and summarizer fail to capture patterns such as room numbering that are critical for efficient exploration.

Scaling to larger maps introduces long-horizon reasoning challenges that degrade performance for all methods. Frontier exploration suffers from high path overhead and large variance, as it does not exploit structural regularities in the layout. NavGPT struggles because its summarizer lacks the detail needed to support long-range planning. LLM Only often loops indefinitely between adjacent cells. Our method succeeds in most cases, but fails when room numbering patterns are ambiguous. For example, in the Large Office environment, room 617 is located near rooms 600A–C, breaking the expected sequential structure.

All methods degrade in the Polycam reconstruction due to geometric artifacts and noise. Our method fails more often because it can become trapped in noisy regions and requires additional steps to recover. However, our method remains more robust as the confidence grid further mitigates noise by downweighting inconsistent predictions over time, allowing the agent to gradually correct false hypotheses as it continues to explore.

Across all settings, Frontier Exploration achieves high success rates but with substantial overhead. LLM Only demonstrates semantic understanding but fails due to poor spatial reasoning, often oscillating between nearby cells. NavGPT is limited by its history buffer and summarizer, which do not capture the structured patterns of real environments. By combining semantic reasoning with frontier-based exploration, our method achieves a balance of high efficiency and robustness, especially in clean structured maps.

However, we observed several failure modes. Our method struggles when numbering patterns are ambiguous (e.g., environments where rooms are labeled 600A, 600B, 617A, 600C in sequence), which produces incorrect semantic hypotheses and biases exploration in the wrong direction. Moreover, when textual labels are sparse or absent, the LLM assigns nearly uniform probabilities across all four regions, causing the confidence grid to converge slowly and reducing overall efficiency.


\subsection{Real-world implementation}

\begin{figure*}[!h]
    \includegraphics[width=\linewidth]{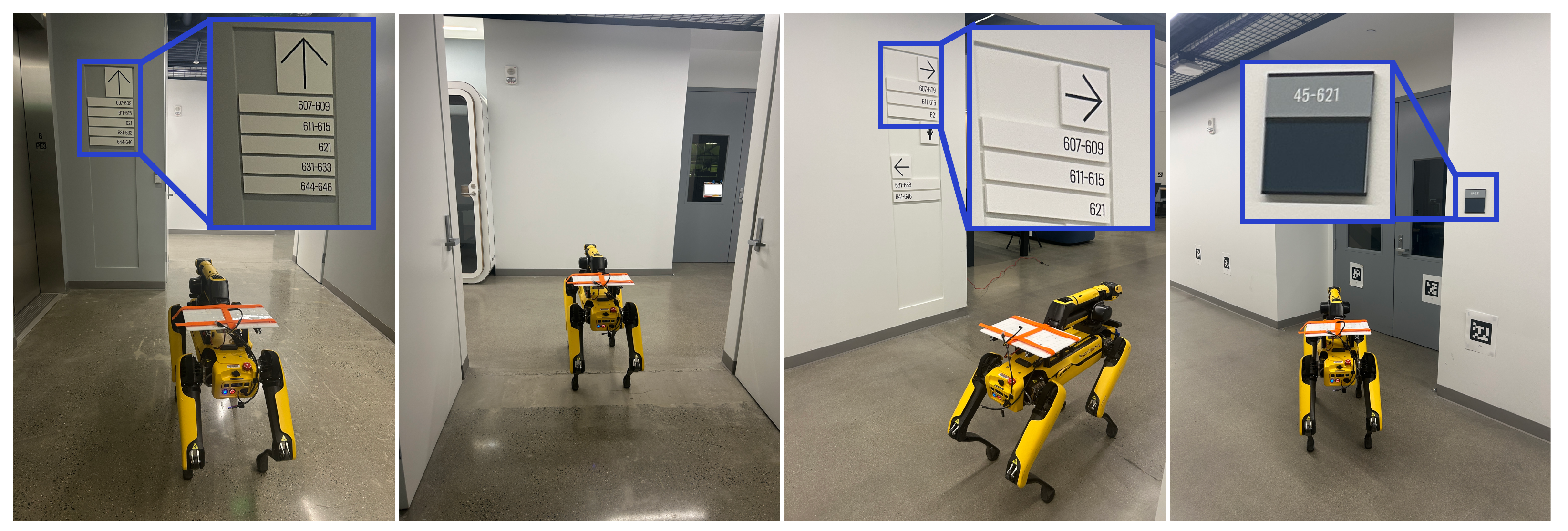}
    \centering
    \caption{Real-world deployment of our method on Spot. The robot is tasked with reaching room 621. Upon encountering an initial sign, Spot infers that room numbers likely increase from left to right and chooses to explore left. It then detects another directional sign confirming the goal’s location and successfully navigates to the target room. Note: the target room was labeled as 621 instead of 45-621 in the ground truth grid world.}
    \label{fig:spot-implementation}
\end{figure*}

To validate our approach in the real world, we deploy it on a Boston Dynamics Spot robot. 
We construct a grid map of the deployment environment with pre-annotated walls, room numbers, and signs in the occupancy and semantic grids, and scale its resolution to match the physical dimensions of the real space. 
The grid is specified with an origin that anchors the grid to the real world and a heading that specifies the orientation of the grid's axes relative to the real world, allowing each cell to be mapped to an SE(2) pose for deployment on the robot.

The signs shown in Figure~\ref{fig:spot-implementation} are explicitly annotated in the semantic grid. The sign in the first image from the left is annotated in the semantic grid as:
"Rooms 607–609, 611–615, 621, 631–633, 644–646 upwards."
The sign in the third image from the left is annotated as:
"Rooms 607–609, 611–615, 621 to the right; Rooms 631–633, 641, 646 to the left."

We calibrate the robot's position in the grid by placing the robot at a known grid cell and computing the fixed transform from the grid frame to Spot's odometry frame and then by combining Spot's current odometry pose with the SE(2) pose of the calibration cell. 
Any subsequent grid cell can then be transformed into the odometry frame and commanded as a navigation goal. 
Motion control is handled via the Boston Dynamics' API to send synchronized SE(2) trajectory points in the odometry frame. 
We monitor command feedback and proceed once the robot is settled at the goal.

At each step of the control loop, Spot updates its visibility grids, queries the LLM for semantic guidance, updates the confidence grid, selects a subgoal, and plans a collision-free path using A*, as described in Section~\ref{sec:method}. The next waypoint and yaw are computed, and Spot advances cell by cell until the goal is reached.

We evaluate the system by starting the robot near the elevator and commanding it to navigate to Room 621. 
The robot successfully reaches the goal, leveraging semantic cues such as hallway signage to bias exploration. 
For example, the first sign encountered near the elevator does not specify whether Room 621 lies to the right or left, yet the system reasons that room numbers typically increase from left to right, and thus begins exploration in the correct direction. A video this exploration process can be seen in the Supplementary materials.

\section{Limitations and Future Work}
\label{sec:conclusion}


This work introduces a hybrid semantic–frontier navigation framework that leverages large language models to infer long-range semantic patterns, such as room numbering schemes and directional signage, and integrates these predictions into a confidence grid to guide exploration. By combining semantic reasoning with classical frontier-based exploration, our method achieves higher SPL than LLM-only or frontier-only baselines across seven environments, ranging from clean synthetic grids to noisy real-world scans. We further validate the approach on a Boston Dynamics Spot robot, where it successfully uses signage cues to navigate to target rooms in a real building, demonstrating the feasibility of bridging language models, perception, and classical planning for structured indoor navigation.

While our results demonstrate the effectiveness of combining LLM-driven semantic reasoning with frontier-based exploration, several limitations remain. In our current setup, room numbers, signs, and other semantic features are explicitly annotated on the grid. The agent also receives full observability of the entire $k \times k$ slice around it, which simplifies perception compared to real-world deployment where visibility is restricted to the robot’s forward-facing sensors. Our method further shows degraded performance in noisy maps, where reconstruction artifacts can trap the agent, and in environments where room numbering patterns are ambiguous or inconsistent.  

Future work will focus on relaxing these assumptions. We plan to integrate image segmentation and vision-language models to enable robust perception of signs, room numbers, and other cues directly from onboard cameras. Online mapping and semantic labeling will be incorporated to eliminate reliance on ground-truth annotations. In addition, we aim to explore uncertainty-aware reasoning to better handle noise and ambiguity, and more scalable memory representations to improve long-horizon performance in large environments.


\bibliographystyle{IEEEtran}
\bibliography{main}

\end{document}